\documentclass[10pt,twocolumn,letterpaper]{article}

\usepackage[pagenumbers]{wacv} 

\usepackage{booktabs}
\usepackage{multirow}
\usepackage[table]{xcolor}

\usepackage{algorithm}
\usepackage{algorithmic}
\usepackage{float}
\usepackage{makecell}

\usepackage{amsmath}
\allowdisplaybreaks

\definecolor{wacvblue}{rgb}{0.21,0.49,0.74}
\usepackage[pagebackref,breaklinks,colorlinks,allcolors=wacvblue]{hyperref}

\def\wacvPaperID{1554} 
\def\confName{WACV}
\def\confYear{2027}

\title{Loss Knows Best: Detecting Annotation Errors in Videos via Loss Trajectories}

\author{
Praditha Alwis \and
Soumyadeep Chandra \and
Deepak Ravikumar \and
Kaushik Roy\\
Purdue University\\
West Lafayette, IN 47907, USA\\
{\tt\small
dalwis@purdue.edu,
chand133@purdue.edu,
dravikum@purdue.edu,
kaushik@purdue.edu}
}

\begin{document}
\maketitle
\begin{abstract}
Reliable video understanding requires high-quality video datasets that can provide both precise semantic labels and temporally consistent annotations. Detecting annotation errors in densely labeled videos is challenging because errors may arise from semantic \textbf{mislabeling}, where labels disagree with visual content, or temporal \textbf{disordering}, where otherwise plausible labels violate procedural progression. Training dynamics have been used to identify mislabeled training examples primarily for static samples. We investigate checkpoint loss dynamics for \textbf{out-of-sample auditing} of temporally annotated videos. We compute \textbf{Cumulative Sample Loss (CSL)} as the mean annotation-conditioned loss of an audit frame across checkpoints trained on a \textit{disjoint} reference set. CSL acts as a dynamic fingerprint and captures the persistent disagreement between its annotation and learned visual-temporal structure. High-CSL frames are then flagged as likely candidates for potential annotation errors, including semantic mislabeling or temporal disordering. Experiments on EgoPER and Cholec80 show that CSL substantially outperforms final-checkpoint loss and achieves up to a \textbf{4.2-point AUC improvement} over prior baselines on EgoPER and \textbf{92.0/78.5 AUC} for mislabeling/disordering on Cholec80. These results demonstrate checkpoint loss dynamics as an effective diagnostic for temporal annotation auditing.

\end{abstract}

\section{Introduction}
\vspace{-0.5em}
High-quality labeled video datasets are essential for learning temporally structured tasks, including action recognition~\cite{slow_fast, TimeSformer, action_segment_1}, phase segmentation~\cite{TeCNO, ASFormer, chandra2025viper}, and procedural understanding~\cite{chowdhury2024opel, chandra2025remap}. Unlike image classification, video annotations must preserve both semantic correctness and temporal consistency, since errors in either dimension degrade downstream model performance. In structured video, errors typically arise either as \textit{semantic mislabeling}, where visually similar steps are confused (e.g., gallbladder retraction versus removal in surgical workflows~\cite{surgical_mislabel}), or as \textit{temporal disordering}, where annotations violate the natural progression of an activity. Because temporal architectures such as Transformers~\cite{attention} and Temporal Convolutional Networks~\cite{surgical_transformer} rely heavily on coherent phase transitions~\cite{surgical_endonet}, unvetted label noise corrupts learned dynamics and destabilizes predictions. With individual videos spanning thousands of densely annotated frames, manual validation is prohibitively expensive, making automated auditing essential-particularly as widely used benchmarks~\cite{athalye, sambasivan} and modern LLM-assisted curation pipelines~\cite{pangakis, wang_LLM} introduce substantial, unverified label noise.


Deep networks exhibit characteristic learning dynamics in which clean and easy examples are often learned before difficult or incorrectly labeled examples~\cite{steinhardt, label_errors_learning_2}. Prior approaches exploit this behavior through training loss, margins, prediction confidence, or their evolution during optimization. Dataset Cartography~\cite{dataset_cartography} characterizes training examples by the confidence and variability across training epochs. Several recent works detect label errors directly from training-loss trajectories~\cite{pleiss2020detecting} or their clustering~\cite{li2025delving, yue2024ctrl}, alongside ensemble- and influence-based approaches such as Confident Learning~\cite{northcutt}.  
A separate line of work, corrective machine unlearning~\cite{goel2024corrective}, aims to remove the influence of corrupted samples from a trained model. Most unlearning methods, however, assume prior knowledge of which samples are corrupted, enabling a binary split into \textit{retain} and \textit{forget} sets. Accurately identifying mislabeled or temporally disordered samples in the first place remains an open problem~\cite{garg2023memorization}, and this challenge is amplified in video, where each sample is a long sequence with dense frame-level annotations, making error localization substantially harder than in static image data.

\begin{figure*}[!ht]
    \centering
    \vspace{-0.8em}
    \includegraphics[width=0.78\linewidth]{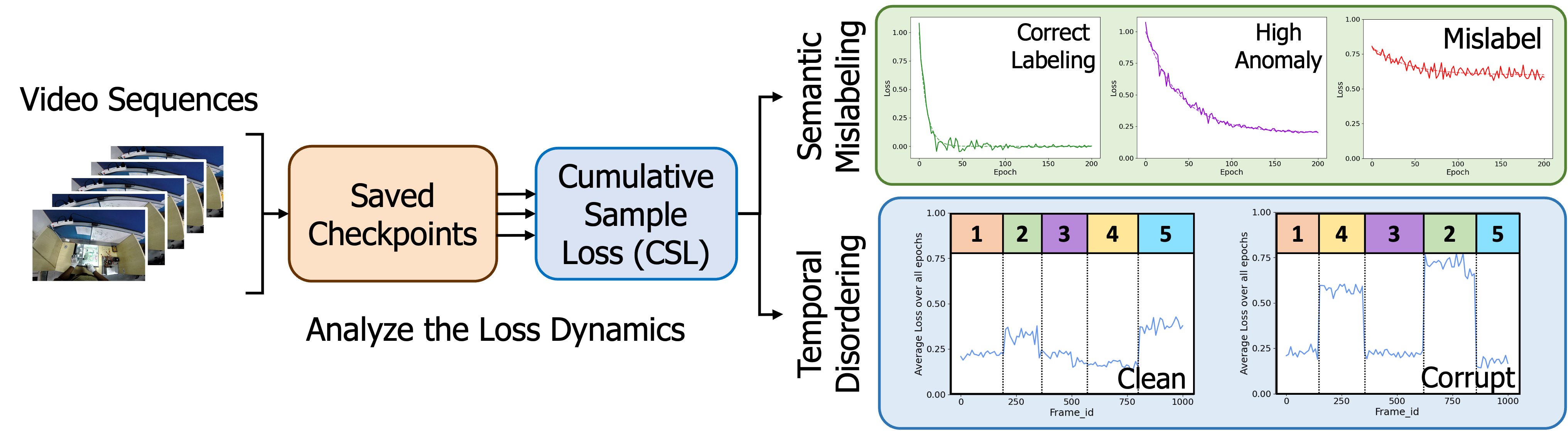}
    \vspace{-0.5em}
    \caption{Overview of our CSL-based framework for detecting annotation errors in video datasets. Top right: frames with correct labels exhibit low and stable sample losses, while mislabeled or ambiguous frames (high anomaly, mislabel) maintain high or erratic loss trajectories. Bottom right: Sequence-level CSL reveals phase disordering. Correctly labeled sequences show smooth loss trends aligned with phase boundaries, whereas corrupted sequences exhibit temporal inconsistencies and abrupt transitions in loss curves.}
    \label{fig:enter-label}
    \vspace{-1.7em}
\end{figure*}


Our question is different: \textit{\textbf{Can training dynamics learned from one set of videos be used to audit annotations in videos that were never used to train the model?}} We study this problem by evaluating a disjoint labeled audit sequence using model checkpoints obtained at different stages of supervised training. Consider an annotated frame whose assigned phase agrees with both its appearance and its temporal context. As the \textbf{reference model} learns the task, increasingly many checkpoints should assign high probability to that annotation. In contrast, an incorrect semantic label can remain inconsistent with the learned visual representation, while a temporally disordered segment can disagree with sequence-level dependencies even if its individual frames remain visually plausible. 

    
We quantify this persistent disagreement using \textbf{Cumulative Sample Loss (CSL)}, defined as the mean annotation-conditioned loss of an audit frame across a collection of training checkpoints. This formulation is related to prior work that characterizes individual samples using model behavior or learning dynamics during training~\cite{dataset_cartography, deepak_angularloss, loss_importance}. The important distinction is the audit setting: the \textit{evaluated sample is not part of the optimization trajectory used to produce the checkpoints}. CSL therefore does not measure how slowly the model memorizes the audited example. Instead, it measures how consistently that annotation disagrees with models representing different stages of learning from an independent reference set. This distinction is particularly useful for video. 
Visual disagreement readily exposes semantic errors, while temporal disordering typically remains hidden until the model leverages sequential context - revealed by sharp spikes in loss at phase boundaries. We therefore systematically analyze the impact of these two corruptions on both frame-level and sequence-aware models via a single, unified signal (Figure~\ref{fig:enter-label}). Our experiments show that this distinction is substantial: frame-wise CNNs are highly effective at identifying semantic mislabeling but perform near chance on temporal disordering, whereas sequence-aware Transformers provide considerably stronger detection of temporal violations.

We evaluate our approach on two temporally structured video benchmarks, \textbf{Cholec80} and \textbf{EgoPER}, spanning surgical workflow analysis and egocentric procedural understanding. On EgoPER, our method achieves frame-level AUC improvements of up to \textbf{4.2 points} over the strongest prior baseline and reaches a mean segment-level error-detection accuracy of \textbf{57.0\%} across five tasks. On Cholec80, our framework reliably achieves \textbf{92.0 AUC} for injected semantic mislabeling and \textbf{78.5 AUC} for phase disordering without prior knowledge of corruption locations.
Our method requires no ground-truth corruption masks, additional supervision, or retraining during auditing. We further analyze its practical tradeoffs through robustness to noisy reference annotations and a cross-validation protocol for extending out-of-sample auditing to an entire dataset.

\vspace{0.2em}
\noindent Our contributions are therefore:
\vspace{0.1em}
\begin{itemize}[leftmargin=*]
    \item We formulate the problem of out-of-sample temporal annotation auditing via training dynamics, distinguishing it from conventional training-sample noise detection based on static loss, confidence, or memorization statistics.
    \item We introduce a lightweight, model-agnostic auditing framework based on \textit{Cumulative Sample Loss (CSL)}, evaluated at inference time from saved checkpoints without auxiliary networks or auditing-specific retraining.
    \item We benchmark our framework on \textbf{Cholec80} and \textbf{EgoPER} against video anomaly- and procedural-error-detection baselines, demonstrating substantially improved error localization over converged single-model baselines.
    \item We provide extensive diagnostic analyses characterizing the impact of backbone architecture, checkpoint density, temporal smoothing, and reference-label corruption on auditing efficacy.
\end{itemize}

\vspace{-0.5em}
\section{Related Work}
\vspace{-0.3em}

\paragraph{Label Noise and Training Dynamics:}
Deep networks fit clean, canonical samples before memorizing noisy or corrupted instances. This small-loss principle has motivated both robust training frameworks, such as co-teaching~\cite{label_errors_learning_2}, and diagnostic tools based on optimization trajectories. For example, \textit{Dataset Cartography}~\cite{dataset_cartography} maps training examples via prediction confidence and variability across epochs, while other approaches leverage forgetting events, margins~\cite{label_errors_learning_5}, influence functions~\cite{label_errors_learning_3}, feature-space clustering~\cite{chen2019understanding,Li_2021_ICCV}, or prediction consensus via \textit{Confident Learning}~\cite{northcutt}. 
Prior methods ~\cite{pleiss2020detecting,ravikumar2025towards} integrate sample loss along the optimization path to suppress single-epoch variance and isolate mislabeled examples. However, these approaches are largely designed for static datasets and may not capture the temporal dependencies underlying video annotations. Our CSL shares this mathematical formulation, connecting it directly to trajectory-based diagnostics. 



\paragraph{Annotation Errors in Temporally Structured Video:}
Assessing annotation quality in video introduces distinct challenges because neighboring frames are strongly correlated and follow task-dependent temporal syntax~\cite{mstcn++,ASFormer,attention,surgical_transformer}. Consequently, label noise is not limited to static \textit{semantic mislabeling}; it frequently manifests as \textit{temporal disordering}, where annotations violate the logical progression of an activity~\cite{label_errors_learning_10}. While static visual mismatch can expose semantic errors, temporal inconsistencies often cannot be detected from individual frame appearances alone. Unlike image-centric noise models or \textit{relaxed} timestamp losses~\cite{label_errors_learning_3}, we explicitly contrast both error modalities and analyze the necessity of temporal sequence modeling for detecting sequence-level violations.

\vspace{-1.4em}
\paragraph{Auditing vs. Robust Training and Unlearning:}
Standard noise-mitigation techniques focus on robust optimization during training (e.g., sample reweighting, pseudo-labeling, noise-transition modeling)~\cite{label_errors_learning_4,label_errors_learning_5,label_errors_learning_6,label_errors_learning_7}. A parallel direction, corrective machine unlearning~\cite{goel2024corrective,kurmanji2023towards,kodge2025sap}, seeks to excise corrupted data post-training. However, unlearning fundamentally presumes that corrupted samples have already been identified, making reliable identification of corrupted samples a critical prerequisite. Our auditing objective is complementary: rather than altering model parameters or modifying training objectives, we focus on identifying and localizing erroneous annotations. 

Crucially, our formulation departs from prior training-dynamics methods in both \textit{where} and \textit{why} the trajectory is evaluated: existing approaches evaluate \textit{in-sample memorization} on data participating directly in parameter updates. In contrast, we use checkpoints from a disjoint reference set to perform \textit{out-of-sample inference-time auditing}, quantifying persistent model-annotation disagreement on unseen video sequences without auxiliary networks, retraining, or ground-truth corruption masks.

\begin{figure*}[!ht]
    \centering
    \includegraphics[width=0.8\linewidth]{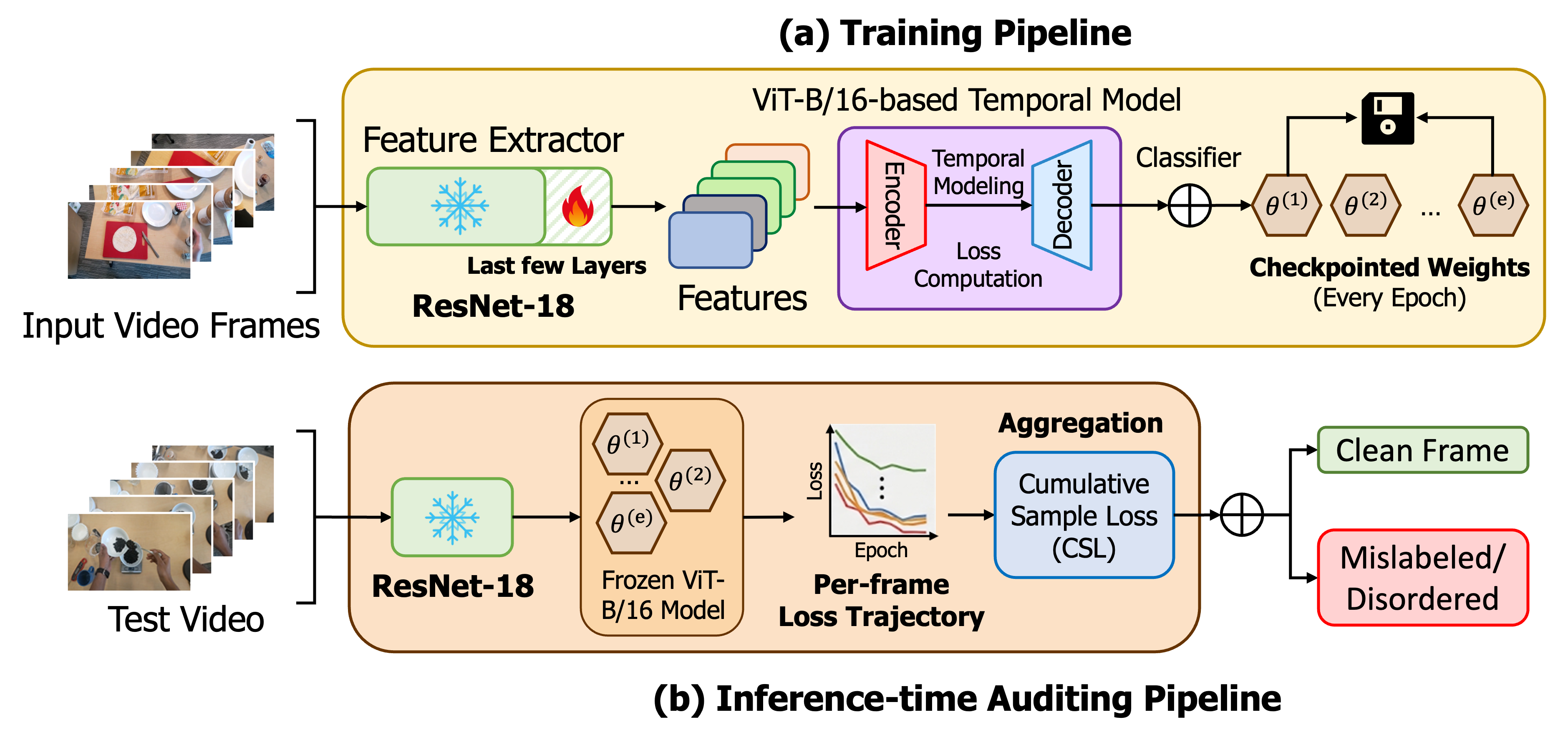}
    \vspace{-1.3em}
    \caption{
    (a) \textbf{Training}: a ResNet-18 feature extractor (FE) produces frame-level features that are passed to a ViT-B/16-based temporal model whose weights $\theta^{(e)}$ are checkpointed at every epoch $e$. (b) \textbf{Inference-time auditing}: each \emph{frozen} checkpoint $\theta^{(e)}$ is applied to the test video with no weight updates; the resulting per-frame loss trajectory across checkpoints is aggregated into the CSL and used to flag mislabeled or disordered frames.}
    \label{fig:pipeline}
    \vspace{-1.6em}
\end{figure*}

\vspace{-0.6em}
\section{Methodology}
\vspace{-0.4em}
\label{sec:method}
We detect annotation errors in temporally labeled video by leveraging \textbf{checkpoint-based loss dynamics}. A model is trained on a training set, with its parameters saved at every epoch. At inference time, each saved checkpoint audits a disjoint labeled test set, producing a loss trajectory across checkpoints for every test frame. Correctly annotated test frames whose assigned phase agrees with both its appearance and its temporal context tends to show decreasing, relatively stable loss as the underlying model becomes better trained. In contrast, mislabeled or temporally inconsistent frames tend to show persistently elevated or unstable loss, reflecting sustained disagreement between model predictions and their assigned annotations. We quantify this behavior using \textbf{CSL}, which aggregates per-frame loss across saved checkpoints into a single annotation-reliability score, computed without additional training on the audited data or prior knowledge of the label-noise distribution.

\vspace{-1em}
\paragraph{Problem Setup:}
Let $\{\mathcal{D_T,D_R}\} = \{(X^{(i)}, Y^{(i)})\}_{i=1}^{N_T,N_R}$ denote a dataset of $N_T$ labeled training videos and disjoint $N_R$ reference videos for auditing. Each video $X^{(i)} = \{x_t^{(i)}\}_{t=1}^{T_i}$ consists of $T_i$ frames, with corresponding frame-level annotations $Y^{(i)} = \{y_t^{(i)}\}_{t=1}^{T_i}$, where $y_t^{(i)} \in \mathcal{Y}$. Our goal is to identify frames with potentially unreliable annotations arising from (i) \textit{semantic mislabeling}, where a frame is assigned an incorrect class or phase label, and (ii) \textit{temporal disordering}, where the assigned labels are locally plausible but violate the expected temporal progression. No corruption labels from $\mathcal D_R$ are used during model training. Unlike training-set loss-dynamics methods, an audit frame does not contribute gradients to any checkpoint used to score it. 

\vspace{-1.2em}
\paragraph{Overview of the Two-Stage Framework:}
\vspace{0.2em}
\begin{enumerate}[leftmargin=*]
    \item \textbf{Training and checkpointing:}
    Train a temporal video model on a training set $D_T$ for $E$ epochs and save the model parameters $\{\theta^{(e)}\}_{e=1}^{E}$.
    \item \textbf{Inference-time auditing:}
    For every video in $D_R$, evaluate each frame using all $E$ saved checkpoints to obtain a checkpoint-based loss trajectory. The losses are aggregated into a CSL score per frame, and high-CSL frames are flagged as potential annotation errors.
\end{enumerate}
Importantly, the auditing stage requires \textbf{no additional training, fine-tuning, or auxiliary models}; it only requires the saved checkpoints from standard model training.

\vspace{-0.2em}
\subsection{Model Architecture}
\label{subsec:architecture}
\vspace{-0.2em}
The CSL computation in Section~\ref{subsec:csl} treats $f_\theta$ as a black box and does not depend on any architectural choice; the only requirement is access to per-epoch checkpoints and a per-frame loss. We instantiate it with the two-stage architecture in Figure~\ref{fig:pipeline}, and separately validate a CNN-only instantiation in Section~\ref{subsec:ablation_temporal} to probe how much of CSL's behavior depends on this specific choice; a broader sweep across architecture families is discussed as future work in the \hyperref[para:limitations]{Limitations} paragraph.

\vspace{-1em}
\paragraph{Feature Extractor:}
We use a ResNet-18 backbone pretrained on ImageNet~\cite{image_net} to extract feature representation from each input frame $x_t$. We freeze early layers to preserve generic visual representations and fine-tune the final layers for domain adaptation.
\vspace{-0.8em}
\begin{equation}
    \mathbf{h}_t = f_{\text{ResNet18}}(x_t),
    \qquad \mathbf{h}_t \in \mathbb{R}^{512}.
    \vspace{-0.8em}
\end{equation}

\vspace{-1em}
\paragraph{Temporal Segmentation:}
Frame-level features are passed through a Transformer-based temporal backbone~\cite{attention} to capture temporal dependencies across the sequence (Figure~\ref{fig:pipeline}(a)). The contextualized representation of frame $t$ is:
\vspace{-0.8em}
\begin{equation}
\mathbf{h}'_t = f_{\text{Transformer}}(\mathbf{h}_{1:T})_t.
\vspace{-0.8em}
\end{equation}

\vspace{-1em}
\paragraph{Classifier Head:}
We map $\mathbf{h}'_t$ to label probabilities with a lightweight MLP head:
\vspace{-0.6em}
\begin{align}
\mathbf{u}_t^{(2)} &=
\sigma\!\left(
\mathrm{LN}\!\left(
W_2\,\sigma\!\left(\mathrm{LN}(W_1\mathbf{h}'_t+b_1)\right)+b_2
\right)\right), \notag\\
\mathbf{z}_t &= W_3\mathbf{u}_t^{(2)}+b_3, \quad
\hat{\mathbf{y}}_t=\mathrm{softmax}(\mathbf{z}_t).
\vspace{-0.8em}
\end{align}
where $W_1\!\in\!\mathbb{R}^{128\times512}$,
$W_2\!\in\!\mathbb{R}^{32\times128}$, and
$W_3\!\in\!\mathbb{R}^{|\mathcal{Y}|\times32}$.
We use ReLU activation, LayerNorm, and dropout rates of 0.5 and 0.3.

\vspace{-0.2em}
\subsection{Training Objective and Checkpoint Saving}
\vspace{-0.2em}
We train the model for $E$ epochs and preserve the model checkpoints $\{\theta^{(1)}, \theta^{(2)}, \dots, \theta^{(E)}\}$ at each training epoch. To handle class imbalance, we use class-weighted cross-entropy. For frame $x_t$ at epoch $e$, the loss is computed as:
\vspace{-0.8em}
\begin{equation}
    \ell_t^{(e)}
    =
    -\sum_{c=1}^{|\mathcal{Y}|}
    \alpha_c \,
    \mathbf{1}[y_t=c]
    \log p_{t,c}^{(e)},
    \vspace{-0.8em}
\end{equation}
where $p_{t,c}^{(e)}$ denotes the predicted probability of class $c$ at epoch $e$, $\mathbf{1}[y_t=c]$ is the class indicator, and $\alpha_c$ is the class-specific weight, chosen inversely proportional to the frequency of class $c$ in the training set.

\vspace{-0.2em}
\subsection{Loss Trajectories and Cumulative Sample Loss}
\label{subsec:csl}
\vspace{-0.2em}
After training, we evaluate each \textbf{test frame} using all saved checkpoints to obtain its per-frame \textbf{loss trajectory}. For a test frame $x_t$ with annotation $y_t$, the loss at checkpoint $\theta^{(e)}$ is:
\vspace{-0.8em}
\begin{equation}
    \hat{\ell}_t^{(e)}
    =
    \mathcal{L}\!\left(
        f_{\theta^{(e)}}(x_t), y_t
    \right),
    \qquad e=1,\ldots,E.
    \vspace{-0.2em}
\end{equation}

We then define the \textbf{Cumulative Sample Loss (CSL)} for each frame $x_t$ as an aggregate per-frame loss trajectory:
\vspace{-0.8em}
\begin{equation}
    \mathrm{CSL}(x_t)
    =
    \frac{1}{E}
    \sum_{e=1}^{E}
    \hat{\ell}_t^{(e)}.
    \vspace{-0.6em}
\end{equation}

As training progresses, later checkpoints generally provide increasingly discriminative representations of the underlying task. We also define a selected checkpoint set $\{\theta^{(1)}, \dots, \theta^{(K)}\}, \; K\le E.$, and discuss the accuracy-compute tradeoff produced by checkpoint subsampling in Supplementary Section~\ref{subsec:checkpoint_sampling}.

CSL serves as a simple summary statistic for \textit{cross-checkpoint disagreement on an unseen annotated sample}. When the annotation agrees with structure learned from the reference data, its loss tends to remain relatively low across sufficiently trained checkpoints. An incorrect annotation can instead remain inconsistent with multiple stages of the learned model, producing a larger integrated disagreement score. A high CSL score does not prove that an annotation is incorrect. Visually ambiguous samples, rare phases, domain shifts, or genuine but unusual executions can also produce elevated loss. We therefore treat CSL as an \textit{auditing score} that ranks annotations for review, rather than as a guaranteed correction mechanism.

\vspace{-1em}
\paragraph{Computational Complexity:}
For a video with $T$ frames and $E$ saved checkpoints, CSL computation requires evaluating the video across all checkpoints, resulting in $\mathcal{O}(ETC_f)$ total computation, where $C_f$ denotes the per-frame inference cost. CSL aggregation adds only $\mathcal{O}(ET)$ operations. Because auditing requires no gradient computation and inference batches and parallelizes efficiently on GPUs, this cost is lower in practice than the asymptotic bound on modern accelerators.

\vspace{-1em}
\paragraph{Smoothing and Sequence-Level Signals:}
Raw CSL values can be noisy due to local visual ambiguity or boundary transitions. We optionally smooth the audit score using a local temporal window:
\vspace{-0.4em}
\begin{equation}
\widetilde{\mathrm{CSL}}(t) = \frac{1}{|\mathcal N_w(t)|}\sum_{j\in\mathcal N_w(t)}\mathrm{CSL}(x_{t+j}),
\vspace{-0.4em}
\label{eqn:csl_smooth}
\end{equation}
where $w$ is a small window size. In practice, smoothing helps reveal temporally coherent error segments: semantic mislabeling often yields sustained high CSL over a contiguous region, while disordering frequently produces sharp spikes around phase transitions.

\begin{algorithm}[!t]
\caption{Annotation Error Detection via CSL}
\label{alg:csl_framework}
\textbf{Inputs:} Training dataset $\mathcal{D}_{T}$, test dataset $\mathcal{D}_{R}$, model $f_{\theta}$, epochs $E$, CSL threshold $\tau$.

\textbf{Output:} Frame-wise anomaly predictions for $\mathcal{D}_{R}$.

\begin{algorithmic}[1]
\STATE Initialize model $f_{\theta^{(0)}}$ (any temporal backbone).
\FOR{epoch $e = 1, 2, \dots, E$}
    \STATE Train $f_{\theta^{(e)}}$ using weighted cross-entropy loss on dataset $\mathcal{D}_{T}$.
    \STATE Save model checkpoint $\theta^{(e)}$.
\ENDFOR

\FOR{video $(X,Y)\in \mathcal{D}_{R}$}
    \FOR{frame $x_t \in X$}
        \STATE Initialize cumulative loss: $ \text{CSL}(x_t) = 0 $
        \FOR{epoch $e = 1, 2, \dots, E$}
            \STATE Compute frame-level checkpoint loss: $\hat{\ell}_t^{(e)} \leftarrow \mathcal{L}(f_{\theta^{(e)}}(x_t), y_t)$.
            \STATE $\mathrm{CSL}(x_t) \leftarrow \mathrm{CSL}(x_t) + \hat{\ell}_t^{(e)}$
        \ENDFOR
        \STATE $\mathrm{CSL}(x_t) \leftarrow \mathrm{CSL}(x_t)/E$
    \ENDFOR
    \STATE Flag errors using threshold $\tau$ or top-$k\%$ ranking. \COMMENT{see Sec.~\ref{subsec:flag}}
    \vspace{-1em}
    \[ \text{Anomaly}(x_t) = \begin{cases} 1, & \text{if } \widetilde{CSL}(x_t) > \tau \\ 0, & \text{otherwise} \end{cases} \]
    \vspace{-0.5em}
\vspace{-1em}
\ENDFOR
\end{algorithmic}
\end{algorithm}

\vspace{-1em}
\subsection{Anomaly Scoring and Error Flagging}
\label{subsec:flag}
We treat both mislabeling and disordering as annotation errors and flag frames using a threshold or percentile rule.

\vspace{-1em}
\paragraph{Threshold-based detection:}
We compute a threshold $\tau$ based on the CSL distribution on a validation set (or robust statistics over the video) and mark frames as anomalous:
\vspace{-0.4em}
\begin{equation}
\mathrm{Anomaly}(x_t)=
\begin{cases}
1, & \text{if } \widetilde{\mathrm{CSL}}(t) > \tau, \\
0, & \text{otherwise.}
\end{cases}
\vspace{-0.8em}
\end{equation}

\vspace{-1em}
\paragraph{Percentile-based detection:}
When tuning $\tau$ is undesirable, we rank frames by $\widetilde{\mathrm{CSL}}(t)$ and flag the top-$k\%$ as candidates. This is convenient for dataset auditing when the goal is to surface the most suspicious regions for review.

Correctly labeled test frames generally exhibit decreasing loss across successive checkpoints, whereas annotation errors maintain elevated loss due to persistent model-annotation disagreement. For semantic mislabeling, high CSL tends to extend across most of the affected segment, reflecting sustained semantic inconsistency (Figure~\ref{fig:csl_maps}). Temporal disordering, in contrast, tends to produce sharp CSL spikes near phase transitions, followed by moderately elevated loss within the disordered segment.

Correctly labeled test frames generally exhibit decreasing loss across successive checkpoints, whereas annotation errors maintain elevated loss due to persistent model--annotation disagreement. For semantic mislabeling, high CSL often extends across most of the affected segment, reflecting sustained semantic inconsistency (Figure~\ref{fig:csl_maps}). In contrast, temporal disordering typically produces sharp CSL spikes near phase transitions, followed by moderately elevated loss within the disordered segment.


\vspace{-0.2em}
\section{Experiments and Results}
\vspace{-0.2em}
We evaluate our framework on two temporally structured video datasets from distinct domains: a surgical workflow dataset (Cholec80) and an egocentric procedural dataset (EgoPER). We assess the model's ability to detect both semantic mislabeling and temporal disordering under controlled annotation corruptions.

\vspace{-0.2em}
\subsection{Dataset}
\vspace{-0.2em}
\textbf{Cholec80}~\cite{surgical_endonet} is a surgical workflow dataset of 80 laparoscopic cholecystectomy videos, annotated at the frame level across seven surgical phases. Following standard protocol, we use 50 videos for training, 10 for validation, and 20 for testing, sampled at 15 fps and resized to $224 \times 224$. To evaluate robustness to annotation errors, we inject two types of synthetic corruption into the test set:
\vspace{0.2em}
\begin{itemize}[leftmargin=2em]
    \item \textbf{Mislabeling:} Continuous frame segments from 10 test videos are assigned incorrect phase labels to simulate semantic annotation errors.
    \item \textbf{Temporal Disordering:} In separate 10 test videos, two contiguous phases are swapped while preserving their frame-level labels, maintaining semantic correctness but violating temporal order.
\end{itemize}
These settings produce two separate evaluation splits, allowing isolated analysis of mislabeling and disordering.

\begin{table}[!ht]
\centering
\vspace{-0.4em}
\caption{Statistics of datasets used. The test sets contain corrupted annotations such as label noise and step disorder.}
\vspace{-0.7em}
\label{tab:dataset}
\resizebox{\linewidth}{!}{%
\begin{tabular}{lccc}
\toprule[1.5pt]
\textbf{Dataset / Task} & \textbf{Train / Test} & \textbf{Duration (min)} & \textbf{Actions} \\
\midrule[1pt]
\multicolumn{4}{l}{\textbf{Cholec80}} \\
\hspace{0.5em}Full Dataset & 50 / 10+20 & 38.0 / 32.3 & 7 \\
\midrule
\multicolumn{4}{l}{\textbf{EgoPER}} \\
\hspace{0.5em}Coffee     & 32 / 35 & 8.4 / 9.3 & 15 \\
\hspace{0.5em}Pinwheels  & 42 / 42 & 5.6 / 4.2 & 13 \\
\hspace{0.5em}Tea        & 47 / 32 & 2.6 / 2.5 & 10 \\
\hspace{0.5em}Quesadilla & 48 / 32 & 1.6 / 1.8 & 8 \\
\hspace{0.5em}Oatmeal    & 44 / 32 & 4.2 / 4.3 & 11 \\
\bottomrule[1.5pt]
\end{tabular}%
}
\vspace{-0.5em}
\end{table}

\begin{table*}[!ht]
\centering
\vspace{-1em}
\caption{
Performance comparison (EDA / AUC) on the \textbf{EgoPER} benchmark. CSL outperforms across all tasks in AUC and achieves competitive EDA, validating its ability to detect diverse annotation errors through cumulative loss dynamics.
}
\vspace{-0.7em}
\label{tab:egoper_results}
\resizebox{\textwidth}{!}{%
\begin{tabular}{lcccccccccc}
\toprule[1.5pt]
\multirow{2}{*}{\textbf{Method}} &
\multicolumn{2}{c}{\textbf{Quesadilla}} &
\multicolumn{2}{c}{\textbf{Oatmeal}} &
\multicolumn{2}{c}{\textbf{Pinwheel}} &
\multicolumn{2}{c}{\textbf{Coffee}} &
\multicolumn{2}{c}{\textbf{Tea}} \\
\cmidrule(lr){2-3}\cmidrule(lr){4-5}\cmidrule(lr){6-7}\cmidrule(lr){8-9}\cmidrule(lr){10-11}
& \textbf{EDA} & \textbf{AUC}
& \textbf{EDA} & \textbf{AUC}
& \textbf{EDA} & \textbf{AUC}
& \textbf{EDA} & \textbf{AUC}
& \textbf{EDA} & \textbf{AUC} \\
\midrule[1pt]
Random                               & 19.9 & 50.0 & 11.8 & 50.0 & 15.7 & 50.0 &  8.2 & 50.0 & 17.0 & 50.0 \\
HF\textsuperscript{2}-VAD~\cite{hf2vad}            & 34.5 & 62.6 & 25.4 & 62.3 & 29.1 & 52.7 & 10.0 & 59.6 & 36.6 & 62.1 \\
HF\textsuperscript{2}-VAD + SSPCAB~\cite{sspcab}   & 30.4 & 60.9 & 25.3 & 61.9 & 33.9 & 51.7 & 10.0 & \textbf{60.1} & 35.4 & 63.2 \\
S3R~\cite{s3r}                                       & 52.6 & 51.8 & 47.8 & 61.6 & 50.5 & 52.4 & 16.3 & 51.0 & 47.8 & 57.9 \\
EgoPED (AF)~\cite{EgoPER}                            & 62.7 & 65.6 & 51.4 & 65.1 & \textbf{59.6} & 55.0 & 55.3 & 58.3 & 56.0 & 66.0 \\
\rowcolor[gray]{0.9}
\textbf{Inference-time CSL (Ours)}                           & \textbf{63.1} & \textbf{68.2} & \textbf{55.4} & \textbf{67.1} & 50.9 & \textbf{57.3} & \textbf{55.8} & 59.6 & \textbf{59.9} & \textbf{70.2} \\
\bottomrule[1.5pt]
\end{tabular}%
}
\vspace{-1.2em}
\end{table*}

\textbf{EgoPER}~\cite{EgoPER} is a large-scale egocentric dataset containing over 28 hours of instructional video captured with a head-mounted GoPro, curated for procedural task understanding and error detection. It spans five tasks: preparing \textit{coffee}, making \textit{tea}, assembling a \textit{pinwheel} sandwich, cooking a \textit{quesadilla}, and making \textit{oatmeal}. Each video is annotated with fine-grained procedural steps and ground-truth transcripts indicating correct and erroneous executions. 

For both datasets, we simulate additional annotation noise, including step reordering, label duplication, and the insertion of incorrect steps, to reflect common inconsistencies in real-world instructional datasets. 



\vspace{-0.3em}
\subsection{Evaluation Metrics}
\label{subsec:metrics}
\vspace{-0.3em}
We evaluate CSL-based annotation error detection using both segment-level and frame-level metrics.

\vspace{-1em}
\paragraph{Segment-wise Error Detection Accuracy (EDA):}
EDA measures the proportion of ground-truth erroneous segments detected within the top-$k\%$ of frames ranked by error score. A segment counts as detected if at least one of its frames is flagged. We compute $\text{EDA} = |\mathcal{D}_e \cap \text{GT}_e| / |\text{GT}_e|$, where $\mathcal{D}_e$ and $\text{GT}_e$ are the predicted and ground-truth erroneous segments.

\vspace{-1em}
\paragraph{Frame-wise Micro-AUC:}
We compute the micro Area Under the ROC Curve (AUC) using frame-level error probabilities, a threshold-independent measure of separability between clean and corrupted frames.

\vspace{-0.2em}
\subsection{Baselines}
\label{subsec:baselines}
\vspace{-0.2em}
We compare our algorithm against video anomaly-detection and procedural-error-detection methods adapted for annotation error detection: HF$^2$-VAD~\cite{hf2vad} and its SSPCAB-enhanced variant~\cite{sspcab}, which rely on reconstruction and future-frame prediction; S3R~\cite{s3r}, which synthesizes pseudo-anomalies via dictionary learning and learns a discriminative decision boundary; and EgoPED~\cite{EgoPER}, which combines action-segmentation backbones (ActionFormer~\cite{ActionFormer}, MSTCN++~\cite{mstcn++}, DiffAct~\cite{diffact}) with graph reasoning and hand-object interaction cues. A random baseline provides a lower bound. Unlike these methods, CSL detects annotation errors from loss dynamics over training checkpoints rather than visual abnormality.

\vspace{-0.2em}
\subsection{Implementation Details}
\label{subsec:implementation}
\vspace{-0.2em}
Our model uses a ResNet-18 backbone for frame-level feature extraction and a ViT-B/16 Transformer for temporal modeling. Models are trained for 200 epochs using AdamW with a learning rate of $10^{-4}$ and class-weighted cross-entropy. Training takes approximately 12 hours on a single NVIDIA A100 GPU.
During inference, CSL is computed per frame by averaging loss across all 200 saved checkpoints, temporally smoothed (Eq.\ref{eqn:csl_smooth}), and frames are ranked to flag errors via fixed percentile thresholds. All hyperparameters are selected via cross-validation. The framework is fully modular and does not require access to corrupted labels during training.

\vspace{-0.2em}
\subsection{Results}
\begin{figure}[!ht]
    \centering
    \vspace{-0.6em}
    \includegraphics[width=\linewidth]{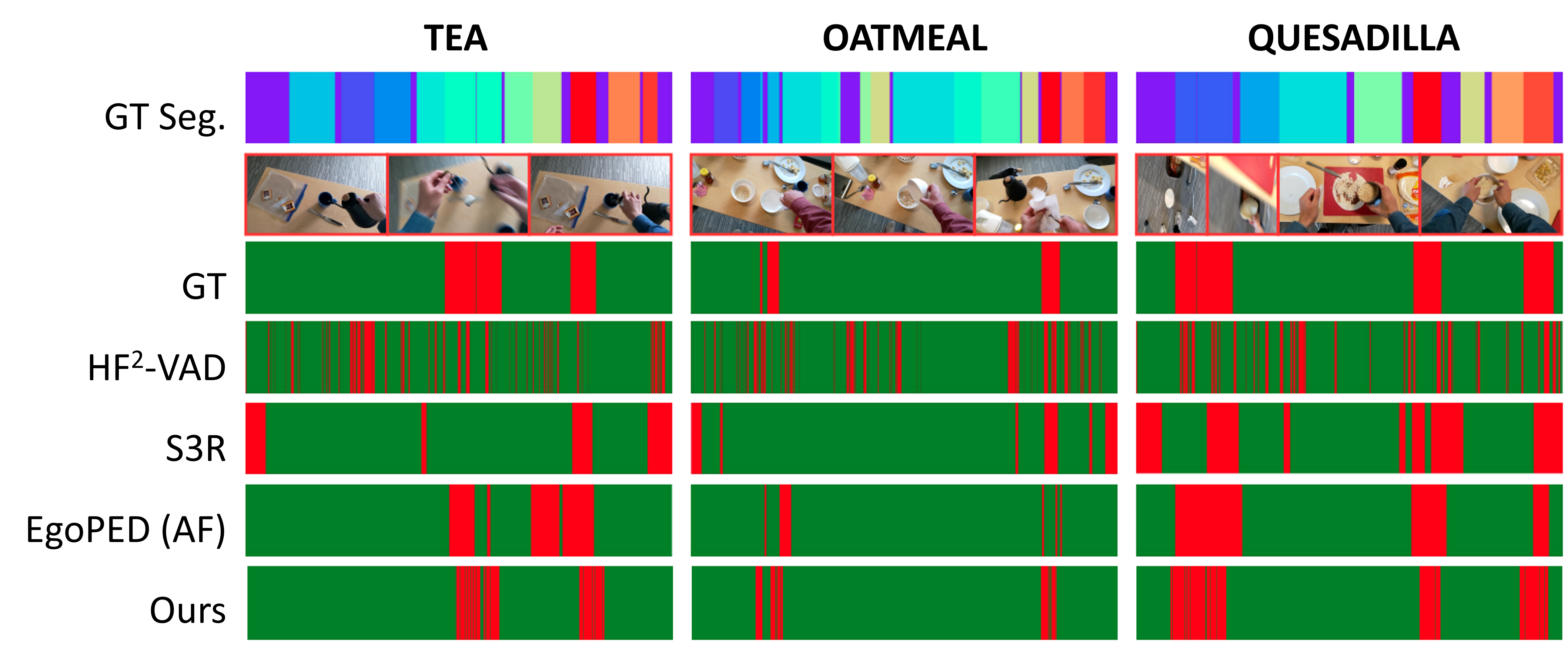}
    \vspace{-2em}
    \caption{
    Error detection across EgoPER test sequences. Green: correct frames. Red: detected annotation errors. Each row is a different method; our method (bottom row) shows precise localization of mislabeled or disordered segments.
    }
    \label{fig:error_visualization_egoper}
    \vspace{-1em}
\end{figure}

\vspace{-0.2em}
\paragraph{EgoPER:}
\label{subsec:egoper_results}
Table~\ref{tab:egoper_results} reports segment-wise EDA and frame-wise AUC across five EgoPER tasks.  CSL achieves the highest AUC on four of five tasks and ties the second-best result on \textit{Coffee}, indicating strong frame-level discrimination between clean and corrupted annotations. It reaches an AUC of \textbf{70.2} on \textit{Tea}, a \textbf{4.2-point} improvement over EgoPED (66.0), and an AUC of \textbf{68.2} on \textit{Quesadilla}, a \textbf{2.6-point} improvement. Averaged across all five tasks, CSL's mean EDA is \textbf{57.0\%}, ranging from 50.9\% (\textit{Pinwheel}) to 63.1\% (\textit{Quesadilla}), demonstrating effective localization of annotation errors at both frame and segment levels. 
Qualitative results in Figure~\ref{fig:error_visualization_egoper} and Supp. S3 show that inference-time CSL produces smoother, more temporally coherent predictions than frame-level anomaly detectors such as HF$^2$-VAD, with fewer isolated false positives.

\begin{figure*}[!ht]
    \centering
    \vspace{-0.4em}
    \includegraphics[width=0.9\linewidth]{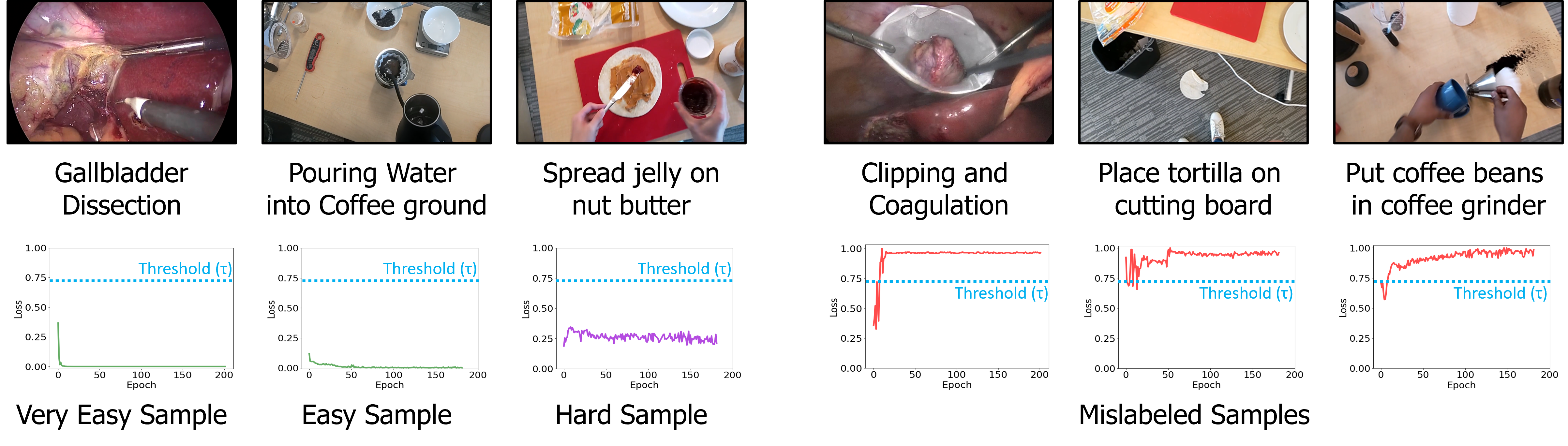}
    \vspace{-0.6em}
    \caption{
    Qualitative results of error detection on EgoPER and Cholec80. Top row: representative frames. Bottom row: cumulative sample loss trajectory over time. Our framework identifies corrupted regions (red spikes) using a task-specific threshold $\tau$.
    }
    \label{fig:error_detection_traj}
    \vspace{-1.5em}
\end{figure*}

\vspace{-1em}
\paragraph{Cholec80:}
\label{subsec:cholec80_results}
Table~\ref{tab:cholec80_results} reports generalization to the Cholec80 surgical dataset under both mislabeling and temporal disordering. Under semantic mislabeling, CSL achieves an EDA of \textbf{85.9} and an AUC of \textbf{92.0}, exceeding the strongest baseline (EgoPED) by \textbf{+19.1} EDA points and \textbf{+20.7} AUC points. In the more challenging temporal disordering setting, where prior baselines do not explicitly report results, CSL reaches an EDA of \textbf{74.5} and an AUC of \textbf{78.5}, demonstrating its ability to capture annotation inconsistencies arising from violations of temporal structure.

\begin{table}[!ht]
\centering
\vspace{-0.6em}
\caption{
Error detection performance (EDA / AUC) on \textbf{Cholec80} under two synthetic annotation-error types: \textit{mislabeling} and \textit{temporal disordering}. Inference-time CSL is the \textit{only method} in this comparison evaluated on both error types.
}
\vspace{-0.5em}
\label{tab:cholec80_results}
\resizebox{\linewidth}{!}{%
\begin{tabular}{ll|cc}
\toprule[1.5pt]
\multicolumn{2}{c|}{\textbf{Method}} & \textbf{EDA} & \textbf{AUC} \\
\midrule[1pt]
Random                             & --          & 24.1 & 50.0 \\
HF\textsuperscript{2}-VAD          & --          & 48.6 & 66.4 \\
HF\textsuperscript{2}-VAD + SSPCAB & --          & 50.3 & 68.1 \\
S3R                                & --          & 57.2 & 63.5 \\
EgoPED (AF)                        & --          & 66.8 & 71.2 \\
\midrule
\multirow{2}{*}{\textbf{Inference-time CSL (Ours)}} 
& \cellcolor[gray]{0.9}\textbf{Mislabeled} 
& \cellcolor[gray]{0.9}\textbf{85.9} 
& \cellcolor[gray]{0.9}\textbf{92.0} \\
& \cellcolor[gray]{0.9}\textbf{Disordered} 
& \cellcolor[gray]{0.9}\textbf{74.5} 
& \cellcolor[gray]{0.9}\textbf{78.5} \\
\bottomrule[1.5pt]
\end{tabular}%
}
\vspace{-1em}
\end{table}

\vspace{-0.2em}
\subsection{Qualitative Analysis}
\label{subsec:qualitative}
\begin{figure*}[!ht]
    \centering
    \vspace{-0.5em}
    \includegraphics[width=0.92\linewidth]{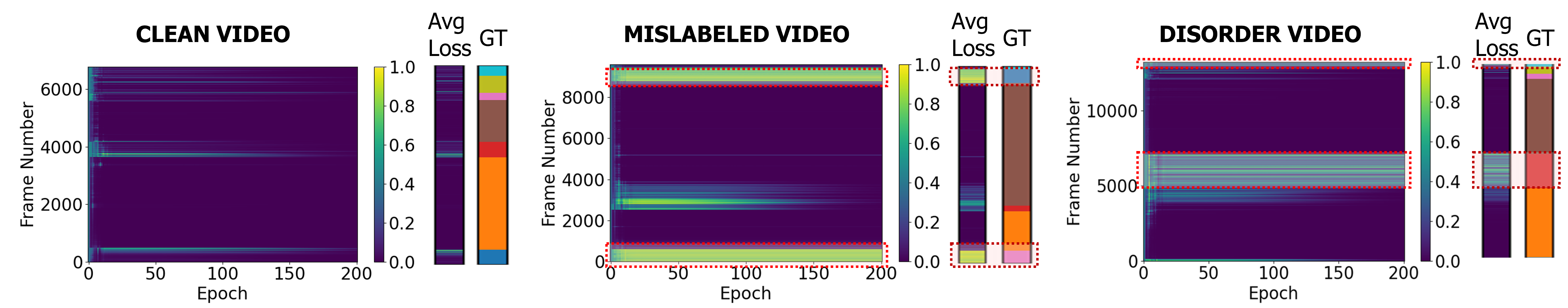}
    \vspace{-0.8em}
    \caption{Visualization of Cumulative Sample Loss (CSL) trends across training epochs. Left: Correctly labeled data exhibits low CSL loss concentration. Center: Label misannotations correspond to high CSL spikes (red boxes). Right: Disordering induces high CSL activations, especially around phase transition boundaries.}
    \label{fig:csl_maps}
    \vspace{-1.4em}
\end{figure*}

\vspace{-0.2em}
Figure~\ref{fig:error_detection_traj}, S1 and S2 visualize CSL trajectories for representative EgoPER and Cholec80 videos. Correctly labeled segments exhibit consistently low CSL, while mislabeled regions produce sustained spikes, and disordered segments show sharp peaks near phase transitions. Figure~\ref{fig:csl_maps} further illustrates these patterns across full sequences, highlighting how CSL naturally separates clean, mislabeled, and temporally inconsistent annotations. These patterns reinforce the effectiveness of our approach in capturing both spatial and temporal annotation inconsistencies through cumulative loss samples. Additional qualitative results are provided in Supp. Fig. S4.

\vspace{-0.3em}
\section{Ablation Studies}
\label{sec:ablation}
\vspace{-0.3em}
We conduct ablation studies to assess the key factors influencing CSL-based annotation error detection. 
We assess three factors: (i) whether checkpoint-based loss dynamics add information beyond final-epoch loss, (ii) how temporal modeling capacity affects detection of ordering errors, and (iii) how CSL behaves when training data itself contains noisy annotations. All ablations use controlled annotation corruption to isolate each factor.

\vspace{-0.2em}
\subsection{Final-Epoch Loss vs. Cumulative Sample Loss}
\vspace{-0.2em}
To evaluate the necessity of checkpoint-based loss dynamics, we compare CSL against loss computed at varying checkpoint densities, ranging from 50\% subsampling down to the final epoch alone. As shown in Table~\ref{tab:final_vs_csl} and Supp. Table~A1, CSL consistently outperforms the single final-epoch baseline across all datasets and corruption types. A converged model provides only a static snapshot of model--annotation agreement, which can obscure errors that become less distinct as optimization proceeds. For semantic mislabeling, late-stage representations may overfit or become overly confident, reducing sensitivity to subtle mismatches. This limitation is even more pronounced for temporal disordering: individual frames often remain visually consistent with their assigned labels at convergence despite violating the temporal sequence. By integrating loss trajectories across the entire training process, CSL captures persistent disagreement throughout optimization, yielding a markedly stronger and more reliable detection signal.

\begin{table}[!ht]
\centering
\vspace{-0.6em}
\caption{AUC comparison between using only the final-epoch loss and the proposed CSL across saved checkpoints.}
\label{tab:final_vs_csl}
\vspace{-0.4em}
\begin{tabular}{ll|c|c}
\toprule[1.5pt]
\multicolumn{2}{c|}{\textbf{Dataset}}
& \shortstack{\textbf{Final-Epoch}\\\textbf{Loss}}
& \textbf{CSL} \\
\midrule[1pt]
Quesadilla & -- & 45.8 & \textbf{68.2} \\
Oatmeal    & -- & 40.1 & \textbf{67.1} \\
Pinwheel   & -- & 45.9 & \textbf{57.3} \\
Coffee     & -- & 42.3 & \textbf{59.6} \\
Tea        & -- & 49.1 & \textbf{70.2} \\
\midrule
\multirow{2}{*}{Cholec80}
& \cellcolor[gray]{0.9}\textbf{Mislabeled}
& \cellcolor[gray]{0.9}81.8
& \cellcolor[gray]{0.9}\textbf{92.0} \\
& \cellcolor[gray]{0.9}\textbf{Disordered}
& \cellcolor[gray]{0.9}44.6
& \cellcolor[gray]{0.9}\textbf{78.5} \\
\bottomrule[1.5pt]
\end{tabular}
\vspace{-0.8em}
\end{table}

\vspace{-0.2em}
\subsection{Temporal Modeling: CNN vs. Transformer}
\vspace{-0.2em}
\label{subsec:ablation_temporal}
We compare a frame-wise CNN against a temporal Transformer on Cholec80 to isolate the role of sequence modeling under identical CSL computation (Table~\ref{tab:CNN_vs_transformers}). This comparison reflects how well each backbone's loss trajectory tracks label reliability.
For temporal disordering, the Transformer substantially outperforms the CNN (\textbf{78.45} vs.\ 48.12 AUC), as its attention mechanisms and positional encodings detect sequence-level violations that the frame-wise CNN cannot capture from local appearance alone. 

\begin{table}[!ht]
\centering
\vspace{-0.6em}
\caption{AUC performance of CNN and Transformer models on mislabeled and disordered Cholec80 test sets.}
\vspace{-0.4em}
\label{tab:CNN_vs_transformers}
\begin{tabular}{l|cc}
\toprule[1.5pt]
\textbf{Dataset} & \textbf{CNN} & \textbf{Transformer} \\
\midrule[1pt]
Cholec80 -- Mislabeled 
& \cellcolor[gray]{0.9}\textbf{98.44} 
& 91.98 \\
Cholec80 -- Disordered 
& 48.12 
& \cellcolor[gray]{0.9}\textbf{78.45} \\
\bottomrule[1.5pt]
\end{tabular}
\vspace{-0.4em}
\end{table}

\noindent Conversely, for semantic mislabeling, the CNN achieves higher accuracy (\textbf{98.44} vs.\ 91.98 AUC), where local visual features suffice, and sequence context mildly dilutes the per-frame error signal, plausibly by smoothing the per-frame loss signal across a temporal window. These findings highlight that frame-wise backbones excel at semantic error detection, whereas sequence-aware architectures are essential for identifying temporal disordering. A practical deployment might therefore ensemble or select between the two by error type rather than defaulting to one architecture.

\vspace{-0.1em}
\subsection{Robustness to Noisy Training Annotations}
\vspace{-0.2em}
\label{subsec:ablation_noise}
Real-world video datasets often contain annotation noise in the \emph{training} set as well as the test set. We corrupt 10\% of the Cholec80 training videos with either semantic \textit{mislabeling} or temporal \textit{disordering}, train on this noisy data, and apply the same inference-time auditing protocol on the corrupted test sets. As Table~\ref{tab:noise_ablation} shows, at this single corruption rate, training noise causes only modest degradation: AUC drops by 0.8pp for mislabeling (92.0 $\rightarrow$ 91.2) and 1.6pp for disordering (78.45 $\rightarrow$ 76.9). 

\begin{table}[!ht]
\centering
\vspace{-0.5em}
\caption{Effect of training annotation noise on AUC performance.}
\vspace{-0.5em}
\label{tab:noise_ablation}
\begin{tabular}{ccc}
\toprule[1.5pt]
\textbf{\shortstack{Cholec\\Training Noise}} &
\textbf{\shortstack{Mislabel\\AUC}} &
\textbf{\shortstack{Disorder\\AUC}} \\
\midrule[1pt]
\rowcolor{gray!30} 0\%  & 92.0 & 78.45 \\
10\% & 91.2 & 76.9  \\
20\% & 90.4 & 75.4  \\
\bottomrule[1.5pt]
\end{tabular}
\vspace{-0.6em}
\end{table}

We attribute this to CSL aggregating loss across the entire training trajectory rather than relying on a single converged model, so that consistently hard-to-learn test frames still produce elevated CSL even when a fraction of training labels are themselves corrupted. 


\vspace{-0.4em}
\section{Limitations}\label{para:limitations}
\vspace{-0.2em}
While out-of-sample loss dynamics provide an effective diagnostic for temporal error detection, our framework has several practical limitations. First, evaluating sequences across saved checkpoints incurs storage and inference overhead that scales linearly with trajectory length. Second, auditing an entire dataset requires $K_f$-fold cross-validation to maintain out-of-sample evaluation, increasing overall training expenditure. Third, elevated CSL reflects persistent model-annotation disagreement rather than definitive label corruption; consequently, rare events, severe visual ambiguity, or distribution shifts can also yield elevated loss. Finally, our quantitative validation focuses on synthetic corruptions across two video benchmarks; evaluating naturally occurring annotation noise (e.g., subtle phase-boundary jitter), broader architecture families (such as MS-TCNs), and multi-rate noise sweeps remains essential future work.

\vspace{-0.2em}
\section{Conclusion}
\vspace{-0.2em}
We present Cumulative Sample Loss (CSL), a framework for detecting annotation errors in temporally labeled video by extending checkpoint-based loss-trajectory analysis, an established signal in the static-data label-noise literature~\cite{dataset_cartography, label_errors_learning_2, northcutt}, to inference-time auditing of held-out test videos, and detection of temporal disordering alongside semantic mislabeling. CSL requires no auxiliary models, handcrafted heuristics, or auditing-specific retraining beyond the checkpoints saved during standard optimization. 
On Cholec80 and EgoPER, CSL reliably detects both semantic mislabeling and temporal disordering under synthetic noise. Our ablations confirm that full checkpoint trajectories outperform final-epoch loss and that sequence-aware backbones are essential for capturing temporal violations. More broadly, our findings substantiate a clear claim: \textit{evaluating training-loss trajectories on held-out video provides an effective, low-supervision diagnostic for auditing both semantic and temporal annotation errors in videos.}

{
    \small
    \bibliographystyle{ieeenat_fullname}
    \bibliography{main}
}

\clearpage
\begin{appendix}
\section{Aggregate Loss Dynamics Across Training Checkpoints}
\label{sec:appendix_loss_dynamics}

\begin{figure*}[!t]
\centering
\includegraphics[width=\linewidth]{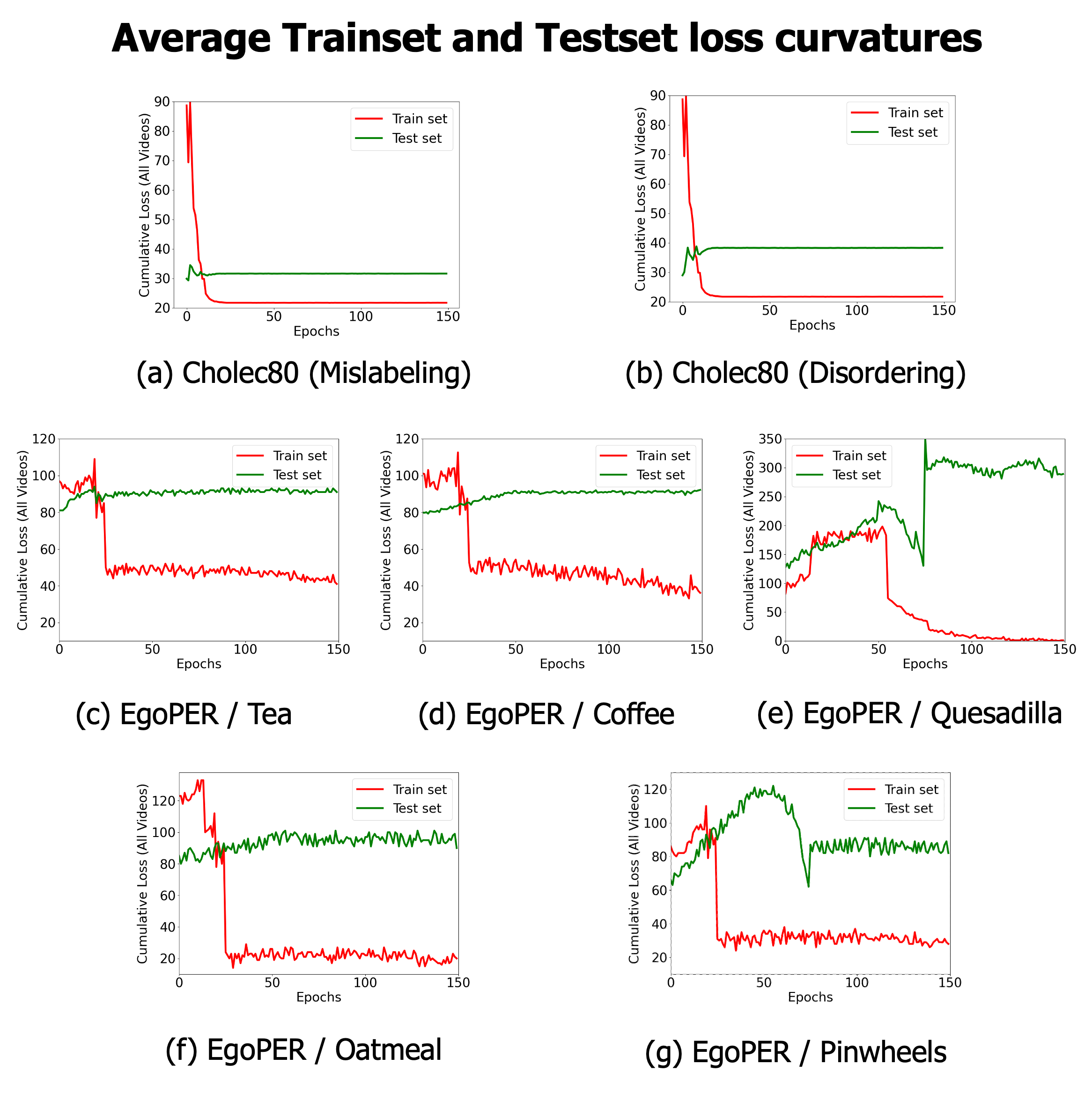}
\caption{Aggregate loss trajectories of the training and held-out test sets across saved checkpoints for Cholec80 and EgoPER tasks. Training loss decreases substantially as the model learns the task, whereas loss evaluated on held-out samples remains comparatively elevated and stabilizes at a distinct level. The resulting separation illustrates how successive checkpoints encode different stages of model--annotation agreement and motivates exploiting the full checkpoint trajectory for post-hoc annotation auditing.}
\label{fig:avg_loss}
\end{figure*}

Figure~\ref{fig:avg_loss} compares aggregate loss trajectories of the training and held-out test sets across saved checkpoints. Across datasets, training loss decreases substantially as optimization progresses, reflecting increasing agreement with the supervision used to train the model. In contrast, loss evaluated on held-out samples follows a distinct trajectory and generally remains elevated after the initial training period. In several EgoPER tasks, this separation becomes particularly pronounced after the early epochs, when training loss rapidly decreases while held-out loss remains high or increases before stabilizing.

Importantly, the held-out samples are never used to update the model. Their trajectories therefore do not describe how these samples are learned or memorized; rather, they measure how models obtained at different stages of training agree with their provided annotations. This distinction motivates our inference-time formulation of CSL. Instead of relying on the prediction behavior of a single converged checkpoint, CSL aggregates model--annotation disagreement across the sequence of learned model states.

The aggregate trends in Figure~\ref{fig:avg_loss} provide dataset-level evidence for this behavior, while the frame-level analyses below show that annotation inconsistencies produce characteristic persistent loss responses within individual sequences.

\section{Frame-Level Checkpoint-Loss Trajectories}
\label{sec:appendix_frame_trajectories}

\begin{figure*}[!t]
    \centering
    \includegraphics[width=0.98\linewidth]{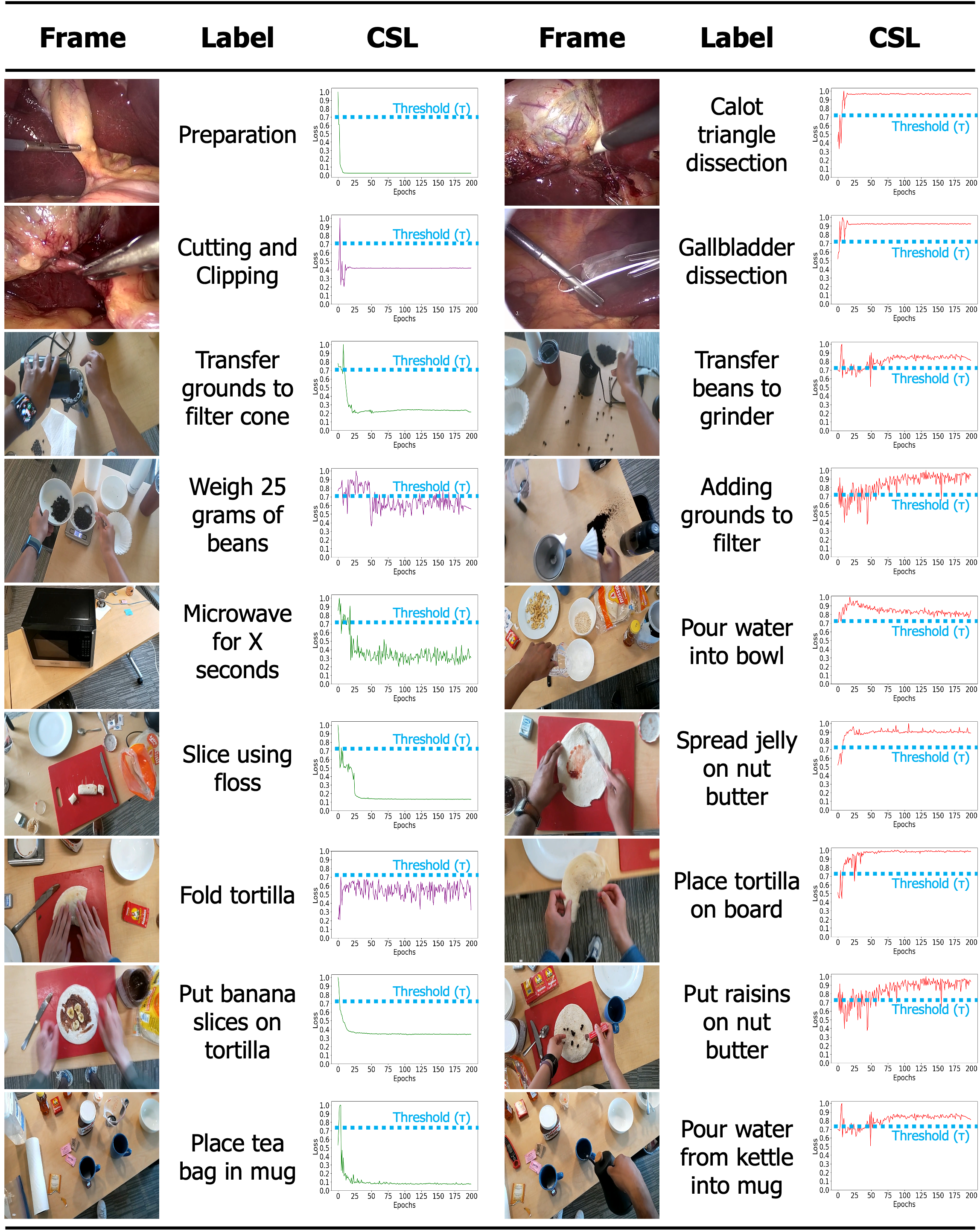}
    \caption{Representative frame-level loss trajectories across saved checkpoints for Cholec80 and EgoPER. Examples with strong model--annotation agreement exhibit low or rapidly decreasing loss, often remaining below the detection threshold $\tau$. In contrast, annotation-inconsistent examples exhibit sustained elevated loss across a large portion of the checkpoint trajectory. These examples illustrate the persistent disagreement summarized by CSL.}
    \label{fig:csl_maps1}
\end{figure*}

Figure~\ref{fig:csl_maps1} provides a frame-level view of the behavior summarized by CSL. For many correctly annotated examples, loss decreases rapidly during the early stages of training and subsequently stabilizes at a low value. Although the evaluated frames are held out from optimization, later checkpoints increasingly agree with their annotations as the model learns the underlying task from the training set.

Annotation-inconsistent examples exhibit a markedly different pattern. Their loss remains elevated across a substantial fraction of the checkpoint trajectory, with several examples consistently exceeding the detection threshold $\tau$. Other examples show substantial early fluctuations before converging to a persistently elevated level. Thus, the relevant signal is not necessarily the loss at any individual checkpoint, but the persistence of disagreement across multiple model states.

This observation provides the intuition behind CSL. A single final-checkpoint loss discards the evolution of model--annotation agreement, whereas CSL integrates evidence across the complete trajectory. Consequently, transient fluctuations contribute less strongly than disagreement that persists over many checkpoints.

The examples also highlight an important limitation: high loss does not by itself prove that an annotation is incorrect. Visually ambiguous or intrinsically challenging frames may also produce elevated loss. CSL should therefore be interpreted as an annotation-auditing score that prioritizes samples for inspection rather than as an absolute measure of annotation correctness.

\section{Qualitative Analysis of Error Localization}
\label{sec:appendix_localization}

\begin{figure*}[t]
    \centering
    \includegraphics[
    width=0.78\linewidth,
    trim={15 0 6 0},
    clip]{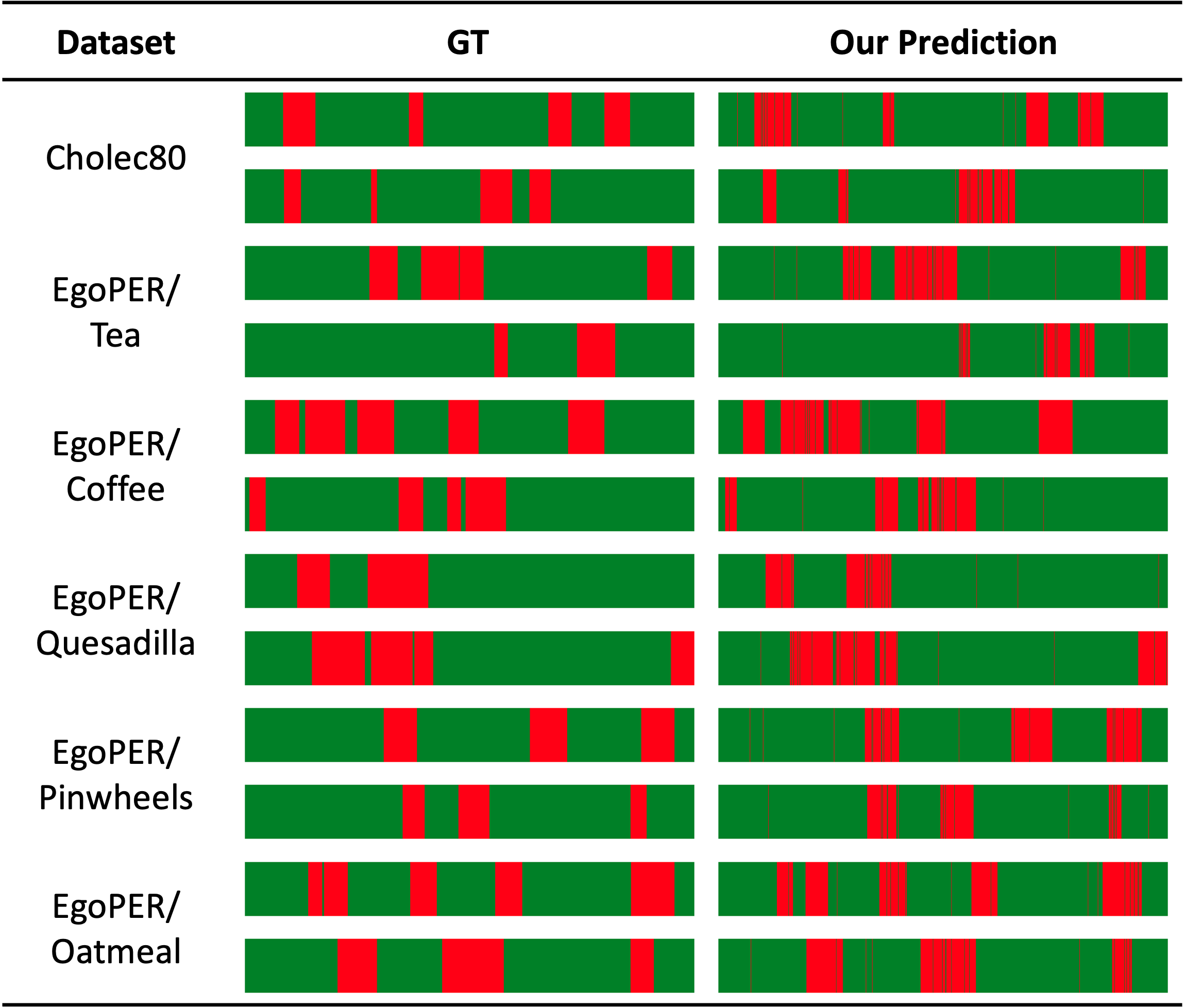}
    \caption{Qualitative comparison of ground-truth and CSL-predicted annotation errors across representative video sequences. Green denotes correctly annotated regions and red denotes annotation errors. CSL recovers both extended corrupted intervals and shorter localized inconsistencies, while occasional fragmented detections occur near error boundaries.}
    \label{fig:qualitative_barplot_comparison}
\end{figure*}

Figure~\ref{fig:qualitative_barplot_comparison} compares ground-truth annotation errors with CSL predictions over complete video sequences. This representation complements frame-level AUC by showing whether high-CSL detections are temporally aligned with the actual corrupted intervals.

Across the illustrated sequences, CSL captures both extended annotation errors and substantially shorter corrupted regions. Large erroneous intervals generally produce contiguous detections with strong temporal overlap, while shorter errors often result in localized responses around the corresponding region. Importantly, the predictions are not uniformly distributed throughout the videos; they are concentrated around temporally coherent regions of annotation inconsistency.

Some predictions are fragmented near error boundaries or extend slightly beyond the ground-truth interval. Such behavior is expected because temporal models incorporate neighboring context, and the transition between consistent and inconsistent annotations may affect predictions on adjacent frames. This also motivates evaluating the framework using complementary metrics: frame-level AUC measures overall discrimination between clean and corrupted frames, whereas segment-level EDA measures whether an erroneous temporal region is successfully identified.

Overall, the sequence-level visualizations show that CSL provides an interpretable temporal auditing signal rather than merely assigning independent anomaly scores to isolated frames.

\section{Spatiotemporal Structure of Checkpoint Loss}
\label{sec:appendix_heatmaps}

\begin{figure*}[!t]
    \centering
    \includegraphics[width=\linewidth]{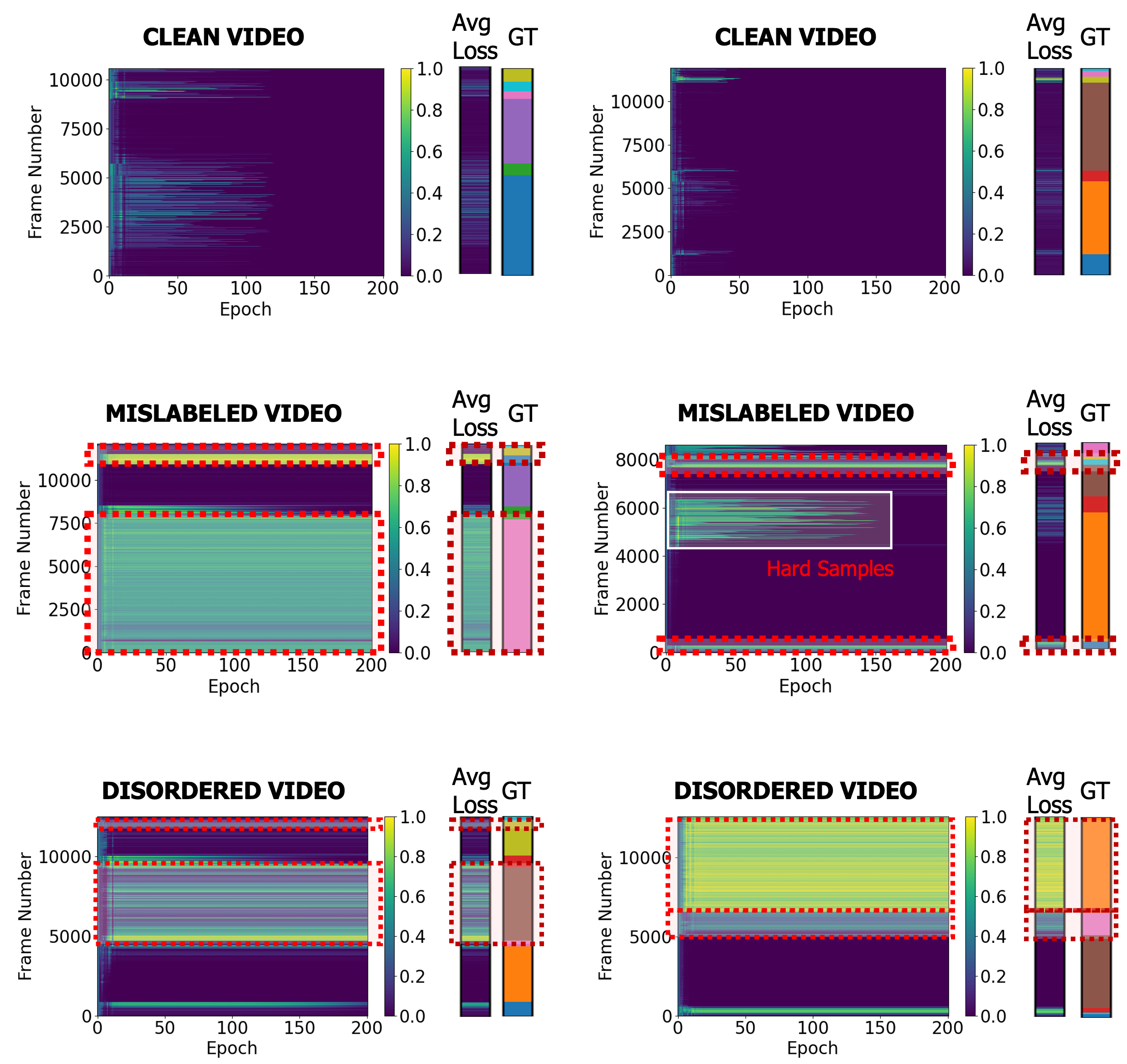}
    \caption{Frame--checkpoint loss heatmaps for representative Cholec80 sequences. Each row corresponds to a frame and each column to a saved training checkpoint; brighter values indicate larger normalized loss. \textbf{Top}: Correctly annotated sequences exhibit predominantly low loss after the early checkpoints. \textbf{Middle}: Localized challenging or corrupted regions form persistent horizontal bands across checkpoints. \textbf{Bottom}: Strong annotation inconsistencies produce broad high-loss regions that persist over a substantial portion of training. Red dashed boxes highlight representative regions of interest.}
    \label{fig:csl_heatmaps_appendix}
\end{figure*}

Figure~\ref{fig:csl_heatmaps_appendix} exposes the two-dimensional structure underlying CSL by visualizing loss jointly over video time (frames) and model training stage (checkpoints). Each horizontal location corresponds to a frame, while movement along the horizontal axis tracks how the loss assigned to that frame changes across successive model checkpoints. CSL can be interpreted as aggregating each frame's trajectory along this checkpoint dimension.

Correctly annotated sequences in the top row exhibit predominantly low loss after the early stages of training. High-loss responses that occur during the initial checkpoints rapidly disappear as the underlying model learns the task. Consequently, these frames accumulate relatively low CSL despite potentially having large loss early in training.

The middle row demonstrates that not every elevated-loss region necessarily corresponds to a clear annotation defect. Certain challenging regions exhibit persistent or slowly decaying loss across multiple checkpoints, as highlighted in the figure. These examples illustrate an important source of ambiguity for loss-based auditing: visually challenging but correctly annotated frames can also produce sustained model disagreement. Nevertheless, their spatial and checkpoint-wise patterns can differ from the broad, strongly elevated responses associated with annotation corruption.

The bottom row shows pronounced persistent loss over large contiguous frame intervals. Unlike transient optimization effects confined to early checkpoints, these regions remain elevated across much of the checkpoint axis. Such persistence indicates that models obtained at substantially different stages of training repeatedly disagree with the provided annotations. Averaging over checkpoints therefore preserves these responses while suppressing short-lived fluctuations.

The heatmaps provide an intuitive explanation for why cumulative checkpoint information is more informative than final-epoch loss alone. Two frames may have similar loss at the final checkpoint despite exhibiting substantially different trajectories during training. CSL retains information about this history of model--annotation agreement, enabling persistent inconsistencies to remain distinguishable even when their final-checkpoint losses are less discriminative.

\section{Additional Ablation Studies}
\label{sec:appendix_ablation}

\subsection{Effect of Checkpoint Sampling Rate}
\label{subsec:checkpoint_sampling}

Evaluating an audit set across all saved checkpoints provides the complete CSL trajectory but increases inference cost. We therefore study whether checkpoint subsampling can reduce computation while preserving detection performance. In addition to Full CSL, we evaluate every 2nd, 5th, and 10th checkpoint and a hybrid strategy that uses every 2nd checkpoint during the first 25\% of training and every 5th checkpoint thereafter.

\begin{table*}[!t]
\renewcommand{\thefigure}{S1}
\centering
\caption{AUC comparison of CSL under different checkpoint sampling strategies.
Final Epoch uses only the last training checkpoint, while Full CSL uses all saved
checkpoints. The $2\times$, $5\times$, and $10\times$ settings compute CSL using
every 2nd, 5th, and 10th checkpoint, respectively. The hybrid strategy samples
every 2nd checkpoint during the first 25\% of training and every 5th checkpoint
during the remaining 75\%.}
\label{tab:csl_checkpoint_sampling}

\resizebox{\textwidth}{!}{%
\begin{tabular}{llcccccc}
\toprule[1.5pt]
\multicolumn{2}{c}{\textbf{Dataset}}
& \shortstack{\textbf{Final}\\\textbf{Epoch}}
& \shortstack{\textbf{Full}\\\textbf{CSL}}
& \shortstack{\textbf{CSL}\\\textbf{($2\times$)}}
& \shortstack{\textbf{CSL}\\\textbf{($5\times$)}}
& \shortstack{\textbf{CSL}\\\textbf{($10\times$)}}
& \shortstack{\textbf{CSL (Hybrid)}\\\textbf{$2\times$ First 25\%, $5\times$ Rest}} \\
\midrule[1pt]

Quesadilla & -- & 45.8 & \textbf{68.2} & 64.3 & 58.7 & 47.1 & 63.8 \\
Oatmeal    & -- & 40.1 & \textbf{67.1} & 62.8 & 56.4 & 42.9 & 62.5 \\
Pinwheel   & -- & 45.9 & \textbf{57.3} & 54.1 & 51.2 & 48.8 & 53.7 \\
Coffee     & -- & 42.3 & \textbf{59.6} & 55.7 & 50.8 & 44.0 & 54.9 \\
Tea        & -- & 49.1 & \textbf{70.2} & 65.8 & 60.7 & 55.5 & 65.6 \\

\midrule

\multirow{2}{*}{Cholec80}
& \cellcolor[gray]{0.9}\textbf{Mislabeled}
& \cellcolor[gray]{0.9}81.8
& \cellcolor[gray]{0.9}\textbf{92.0}
& \cellcolor[gray]{0.9}89.1
& \cellcolor[gray]{0.9}87.0
& \cellcolor[gray]{0.9}85.7
& \cellcolor[gray]{0.9}88.5 \\

& \cellcolor[gray]{0.9}\textbf{Disordered}
& \cellcolor[gray]{0.9}44.6
& \cellcolor[gray]{0.9}\textbf{78.5}
& \cellcolor[gray]{0.9}70.8
& \cellcolor[gray]{0.9}62.4
& \cellcolor[gray]{0.9}52.3
& \cellcolor[gray]{0.9}69.7 \\

\bottomrule[1.5pt]
\end{tabular}%
}

\end{table*}

Table~\ref{tab:csl_checkpoint_sampling} shows a consistent relationship between checkpoint density and annotation error detection performance. Full CSL achieves the highest AUC across all EgoPER tasks and both Cholec80 corruption settings, confirming that intermediate checkpoints contain useful information beyond the final model state.

Moderate subsampling preserves much of this benefit for semantic mislabeling. On Cholec80, using every second checkpoint achieves \textbf{89.1} AUC compared with \textbf{92.0} for Full CSL, while even $10\times$ sampling retains \textbf{85.7} AUC. In contrast, temporal disordering is substantially more sensitive to checkpoint density: AUC decreases from \textbf{78.5} with Full CSL to \textbf{70.8}, \textbf{62.4}, and \textbf{52.3} under $2\times$, $5\times$, and $10\times$ sampling, respectively.

This difference is consistent with the qualitative dynamics in Figure~\ref{fig:csl_heatmaps_appendix}. Semantic mislabeling often produces sustained disagreement over long portions of the checkpoint trajectory, allowing the signal to remain observable under moderate subsampling. Temporal inconsistencies can exhibit more checkpoint-dependent responses, making dense sampling more important for preserving their trajectory structure.

The hybrid strategy provides an intermediate computation--accuracy trade-off, achieving \textbf{88.5} AUC for mislabeling and \textbf{69.7} for disordering on Cholec80 while requiring fewer checkpoint evaluations than Full CSL. These results suggest that checkpoint subsampling is feasible when computational resources are limited, although aggressive subsampling can substantially reduce sensitivity to temporal annotation errors.

\subsection{Effect of Feature Extractor Adaptation}
\label{subsec:ablation_feature}

We further investigate how the quality of the underlying visual representation affects CSL by comparing a fully frozen ResNet-18 feature extractor with a partially fine-tuned variant in which the final two layers are trainable.

\begin{table}[!t]
\centering
\vspace{-0.5em}
\renewcommand{\thetable}{A2}
\caption{AUC values of frozen and trainable ResNet18 across datasets.}
\label{tab:train_vs_frozen}
\begin{tabular}{llcc}
\toprule[1.5pt]
\multicolumn{2}{c}{\textbf{Dataset}} 
& \shortstack{\textbf{Frozen}\\\textbf{ResNet18}} 
& \shortstack{\textbf{Trainable}\\\textbf{ResNet18}} \\
\midrule[1pt]
Quesadilla & -- & 46.2 & 68.2 \\
Oatmeal    & -- & 41.4 & 67.1 \\
Pinwheel   & -- & 40.9 & 57.3 \\
Coffee     & -- & 36.7 & 59.6 \\
Tea        & -- & 48.1 & 70.2 \\
\midrule
\multirow{2}{*}{Cholec80} 
& \cellcolor[gray]{0.9}\textbf{Mislabeled} & \cellcolor[gray]{0.9}\textbf{88.7} & \cellcolor[gray]{0.9}\textbf{92.0} \\
& \cellcolor[gray]{0.9}\textbf{Disordered} & \cellcolor[gray]{0.9}\textbf{76.6} & \cellcolor[gray]{0.9}\textbf{78.5} \\
\bottomrule[1.5pt]
\end{tabular}
\vspace{-1em}
\end{table}

Table~\ref{tab:train_vs_frozen} shows that partial fine-tuning consistently improves annotation error detection. The effect is particularly pronounced on EgoPER: AUC increases from \textbf{46.2} to \textbf{68.2} on Quesadilla, \textbf{41.4} to \textbf{67.1} on Oatmeal, and \textbf{36.7} to \textbf{59.6} on Coffee. These gains indicate that generic ImageNet features alone may not sufficiently represent the fine-grained visual cues required for procedural video understanding.

The improvement can be explained through the model--annotation disagreement measured by CSL. When the underlying representation is poorly aligned with the target domain, even correctly annotated audit samples may receive elevated loss because the model cannot reliably represent the relevant visual evidence. This reduces separation between clean and corrupted annotations. Partial fine-tuning improves task-specific representations, resulting in more informative checkpoint-loss trajectories and stronger separation.

The effect is smaller on Cholec80, where fine-tuning improves AUC from \textbf{88.7} to \textbf{92.0} for mislabeling and from \textbf{76.6} to \textbf{78.5} for disordering. This suggests that the pretrained representation already provides useful visual features for this domain, while task-specific adaptation offers an additional improvement.

Importantly, these results clarify the meaning of model agnosticism in our framework. CSL does not require a specific architecture, feature extractor, or temporal backbone; however, its auditing signal necessarily depends on the underlying model learning meaningful task structure. CSL is therefore \emph{model-agnostic but not model-independent}: improvements in the underlying representation can directly improve the quality of the resulting annotation-auditing signal.
\end{appendix}

\end{document}